\documentclass[10pt,twocolumn,letterpaper]{article}

\usepackage{cvpr}
\usepackage{times}
\usepackage{epsfig}
\usepackage{graphicx}
\usepackage{amsmath}
\usepackage{amssymb}
\usepackage{multirow}
\usepackage{arydshln}

\cvprfinalcopy 

\def\cvprPaperID{3652} 

\ifcvprfinal\fi
\begin{document}

\title{Progressive Attention Memory Network for Movie Story Question Answering}


\author{Junyeong Kim$^1$\thanks{This research was supported by Samsung Research}\ \ \ \ \ \ Minuk Ma$^1$\ \ \ \ \ \ Kyungsu Kim$^2$\ \ \ \ \ \ Sungjin Kim$^2$\ \ \ \ \ \ Chang D. Yoo$^1$ \\$^1$ Korea Advanced Institute of Science and Technology (KAIST) \\ $^2$ Samsung Research \\{\tt\small $^1$\{junyeong.kim, akalsdnr, cd\_yoo\}@kaist.ac.kr}\ \ \ {\tt\small $^2$\{ks0326.kim, sj9373.kim\}@samsung.com}}

\maketitle

\begin{abstract}
   This paper proposes the progressive attention memory network (PAMN) for movie story question answering (QA). Movie story QA is challenging compared to VQA in two aspects: (1) pinpointing the temporal parts relevant to answer the question is difficult as the movies are typically longer than an hour, (2) it has both video and subtitle where different questions require different modality to infer the answer. To overcome these challenges, PAMN involves three main features: (1) progressive attention mechanism that utilizes cues from both question and answer to progressively prune out irrelevant temporal parts in memory, (2) dynamic modality fusion that adaptively determines the contribution of each modality for answering the current question, and (3) belief correction answering scheme that successively corrects the prediction score on each candidate answer. Experiments on publicly available benchmark datasets, MovieQA and TVQA, demonstrate that each feature contributes to our movie story QA architecture, PAMN, and improves performance to achieve the state-of-the-art result. Qualitative analysis by visualizing the inference mechanism of PAMN is also provided.
\end{abstract}

\section{Introduction}
\label{sec:1}
Humans have an innate cognitive ability to infer from different sensory inputs to answer questions of 5W's and 1H involving \textit{who, what, when, where, why} and \textit{how}, and it has been a quest of mankind to duplicate this ability on machines. In recent years, studies on question answering (QA) have successfully benefited from deep neural networks, and showed remarkable performance improvement on textQA \cite{e2ememory,memory}, imageQA \cite{Anderson2017up-down,antol2015vqa,malinowski2015ask,SAN}, videoQA \cite{motion-appearance,jang-CVPR-2017,Yu_2018_ECCV,zhu2017uncovering}. This paper considers movie story QA \cite{mdam,fvta,rwmn,MovieQA,lmn} that aims at a joint understanding of vision and language by answering questions about movie contents and storyline after observing temporally-aligned video and subtitle. Movie story QA is challenging compared to VQA in following two aspects: (1) pinpointing the temporal parts relevant to answer the question is difficult as the movies are typically longer than an hour and (2) it has both video and subtitle where different questions require different modality to infer the answer.

The first challenge of movie story QA is that it involves long videos that are possibly longer than an hour which hinders pinpointing the required temporal parts. The information in the movie required to answer the question is not distributed uniformly across the temporal axis. To address this issue, memory networks \cite{e2ememory} have widely been accepted in QA tasks \cite{rwmn,e2ememory,MovieQA,memory}. The attention mechanism have widely been adopted to retrieve the information relevant to the question. We observed that single-step attention on memory networks \cite{rwmn,MovieQA} often generates blurred temporal attention map.

The second challenge of movie story QA is that it involves both video and subtitle where different questions require different modality to infer the answer. Each modality may convey essential information for different questions, and optimally fusing them is an important problem. For example in the movie \textit{Indiana Jones and the Last Crusade}, answering the question \textit{\textquotedblleft What does Indy do to the grave robbers at the beginning of the movie?\textquotedblright} would require video modality rather than subtitle modality while the question \textit{\textquotedblleft How has the guard managed to stay alive for 700 years?\textquotedblright} would require subtitle modality. Existing multi-modal fusion methods \cite{mcb,MLB,mdam} only focus on modeling rich interactions between the modalities. However, these methods are question-agnostic in that the fusion process is not conditioned on the question.

To address the aforementioned challenges, this paper proposes Progressive Attention Memory Network (PAMN) for movie story QA. PAMN contains three main features; (1) progressive attention mechanism for pinpointing required temporal parts, (2) dynamic modality fusion for adaptively fusing modalities conditioned on question and (3) belief correction answering scheme. Progressive attention mechanism utilizes cues from both question and answers to prune out irrelevant temporal parts for each memory. While iteratively taking question and answers for temporal attention generation, memories are progressively updated to accumulate cues to locate relevant temporal parts for answering the question. Compared to stacked attention \cite{Fan_2018_CVPR,SAN}, progressive attention considers multiple sources (e.g., Q and A) and multiple targets (e.g., video and subtitle memory) in a single framework. 
Dynamic modality fusion aggregates the outputs from each memory by adaptively determining the contribution of each modality. Conditioned on the current question, the contribution is obtained by soft-attention mechanism. Fusing multi-modal data by bilinear operations \cite{mutan,mcb, MLB} often requires heavy computation or large number of parameters. Dynamic modality fusion efficiently integrate video and subtitle modality by discarding worthless information from unnecessary modality.
Belief correction answering scheme successively corrects the prediction score of each candidate answer. When humans solve questions, they typically read content, question and answers multiple times in an iterative manner \cite{human}. This observation is modeled by belief correction answering scheme. The prediction score (logits), which this paper refers to a belief, is equally likely initialized and successively corrected compared to existing answering scheme \cite{mdam,rwmn,lmn} which uses single-step answering scheme.

The main contribution of this paper is summarized as follows. (1) This paper proposes a movie story QA architecture referred to as PAMN that tackles major challenges of movie story QA with three features; progressive attention, dynamic modality fusion and belief correction answering scheme. (2) PAMN achieves the state-of-the-art results on MovieQA dataset. Both the quantitative and qualitative results exhibit the benefits and potential of PAMN.

\section{Related Work}
\label{sec:2}

\subsection{Visual Question Answering}
\label{ssec:2.1}

Despite the short history, imageQA enjoys large number of datasets including VQA~\cite{antol2015vqa}, COCO-QA~\cite{ren2015image} and Visual7W~\cite{zhu2016cvpr}. Attention mechanism is widely used to locate the visual clues relevant to the question. Stacked Attention Network (SAN)~\cite{SAN} utilizes stacked attention module to query an image multiple times to infer the answer progressively. The Dual Attention Network (DAN)~\cite{DAN} jointly leverages visual and textual attention mechanisms to localize key information from both image and question. Recently, applying bilinear operation showed promising results on imageQA. Multimodal Compact Bilinear pooling (MCB)~\cite{mcb} utilized bilinear operation to fuse image and question features in imageQA. To reduce the computational complexity, MCB uses the sampling-based approximation. To further reduce the feature dimension, Multimodal Low-rank Bilinear Attention Network (MLB)~\cite{MLB} utilizes Hadamard product in the common space with two low-rank projection matrices. Multimodal Tucker Fusion \cite{mutan} utilizes tucker decomposition \cite{Tucker1966} to efficiently parameterize bilinear interactions between visual and textual representation.

VideoQA is a natural extension of imageQA as video can be seen as temporal extension of image. Large-scale videoQA benchmarks such as TGIF-QA~\cite{jang-CVPR-2017} and \textquoteleft fill-in-the-blank\textquoteright~\cite{zhu2017uncovering} have boosted the research on videoQA. Spatio-temporal VQA (ST-VQA)~\cite{jang-CVPR-2017} generates spatial and temporal attention to localize which regions in a frame and which frames in a video to attend, respectively. Yu \textit{et al.} \cite{Yu_2018_ECCV} proposed Joint Sequence Fusion (JSFusion) that measures semantic similarity between video and language. JSFusion utilizes hierarchical attention mechanism that learns matching representation patterns between modalities.

\subsection{Movie Story Question Answering}
\label{ssec:2.2}
A recent direction in videoQA leverages text modality such as subtitle in addition to video modality for story understanding. To this end, various video story QA benchmarks such as PororoQA \cite{demn}, MeMexQA~\cite{jiang2017memexqa}, TVQA \cite{lei2018tvqa} and MovieQA~\cite{MovieQA} have been suggested. Numerous researches have tackled MovieQA benchmark which provides movie clip, subtitle and other various textual descriptions. Tapaswi \textit{et al.}~\cite{MovieQA} divided the movie into multiple sub-shots and utilized memory network (MemN2N) \cite{e2ememory} to store video and subtitle features into memory slots. Deep Embedded Memory Network (DEMN) ~\cite{demn} reconstructs stories from a joint stream of scene-dialogue using a latent embedding space and retrieves information which is relevant to the question. Na \textit{et al.} \cite{rwmn} proposed Read-Write Memory Network (RWMN) which is a CNN-based memory network where video and subtitle features are first fused using bilinear operation, then write/read networks store/retrieve information, respectively. 

Liang \textit{et al.} \cite{fvta} proposed Focal Visual-Text Attention (FVTA) that utilizes the hierarchical attention applied to a three-dimensional tensor to localize evidential image and text snippets. Layered Memory Network (LMN)~\cite{lmn} utilizes Static Word Memory module and Dynamic Subtitle Memory module to learn frame-level and clip-level representations. The hierarchically formed movie representation encodes the correspondence between words and frames, and the temporal alignment between sentences and frames. Multimodal Dual Attention Memory (MDAM) \cite{mdam} utilizes multi-head attention mechanism \cite{NIPS2017_7181} and question attention to learn the latent concepts of multimodal contents. Multimodal fusion is performed once after the attention process. Compared to existing architectures on movie story QA that adopt single-step reasoning, PAMN provides multi-step reasoning approach to localize necessary information from question, answers, and movie contents.

\begin{figure*}[t!]
	\begin{center}
		\includegraphics[width=\textwidth]{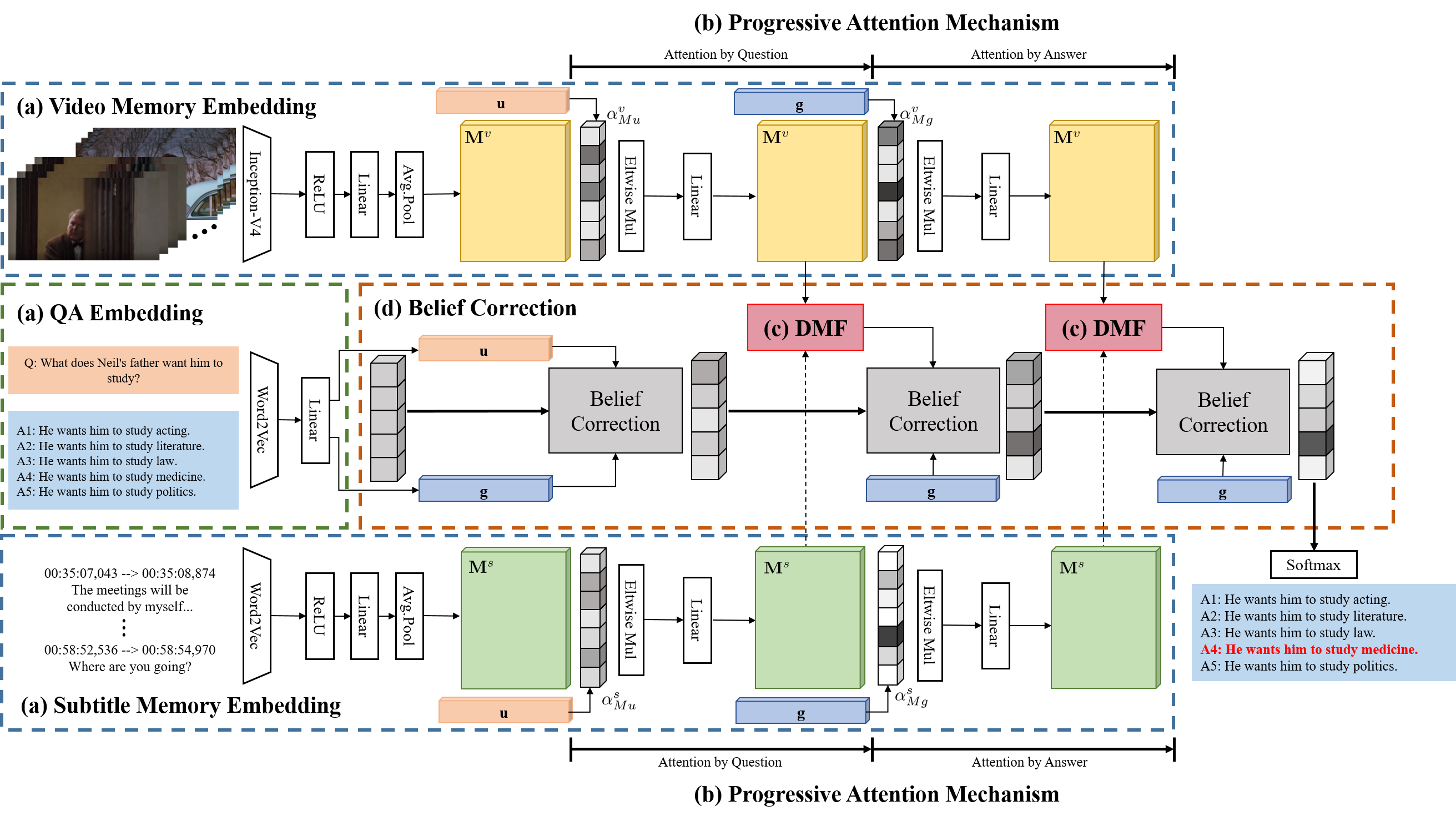}
		\caption{Illustration of the proposed PAMN. The pipeline of PAMN is as follows. (a) Question and candidate answers are embedded into a common space. Video and subtitle are embedded into dual memory that holds independent memories for each modality. (b) Progressive attention mechanism pinpoints temporal parts that are relevant to answering the question. To infer the correct answer, (c) dynamic modality fusion that adaptively integrates outputs of each memory by considering contribution of each modality. (d) Belief correction answering scheme successively corrects the belief of each answer from equally likely initialized belief.}
		\label{fig:overall}
	\end{center}
\end{figure*}

\section{Progressive Attention Memory Network}
\label{sec:3}

This section describes the proposed Progressive Attention Memory Network (PAMN). Fig. \ref{fig:overall} shows the overall architecture of PAMN,  which fully utilizes diverse sources of information (video, subtitle, question and candidate answers) to answer the question. The pipeline of PAMN is as follows. First, video and subtitle are embedded into dual memory as in Fig. \ref{fig:overall}(a) that holds independent memories for each modality. Then, progressive attention mechanism pinpoints temporal parts that are relevant to answering the question as in Fig. \ref{fig:overall}(b). To infer the correct answer, dynamic modality fusion in Fig. \ref{fig:overall}(c) adaptively integrates outputs of each memory by considering contribution of each modality. Belief correction answering scheme successively corrects the belief of each answer from equally likely initialized belief as in Fig. \ref{fig:overall}(d).

\subsection{Problem Setup}
\label{ssec:3.1}
The formal definition of the problem is as follows. The inputs of PAMN are (1) a question representation $\textbf{q} \in \mathbb{R}^{300}$, (2) five candidate answer representations $\{\textbf{a}_i\}_{i = 1}^{5} \in \mathbb{R}^{5 \times 300}$, (3) temporally aligned video ($\textbf{v}$) and subtitle ($\textbf{s}$) representation $\{(\textbf{v}_i, \textbf{s}_i)\}_{i=1}^{T}$ on the whole movie. Each element of subtitle representation $\textbf{s}_i$ corresponds to the dialog sentence of a character and each element of video representation $\textbf{v}_i$ is extracted from temporally aligned video clip. The number of overall sentences of the movie is denoted as $T$. The detailed explanation on extracting visual and textual feature is provided in Section \ref{ssec:4.2}. The objective is to maximize the following likelihood:
\begin{equation}
\text{arg max}_\theta\sum_{\mathcal{D}}\log P(\mathbf{y} | \mathbf{v, s, q, a};\boldsymbol{\theta}),
\end{equation}
where $\boldsymbol{\theta}$ denotes learnable model parameters, $\mathcal{D}$ represents dataset and $\mathbf{y}$ represents the correct answer.

\subsection{Dual Memory Embedding}
\label{ssec:3.2}
As depicted in Fig. \ref{fig:overall}(a), the inputs are first mapped to an embedding space. The question representation $\textbf{q}$ and candidate answer representations $\{\textbf{a}_i\}_{i = 1}^{5}$ are embedded to a common space by weight-shared linear fully connected (FC) layer with parameters $\mathbf{W}_{ug} \in \mathbb{R}^{300 \times d}$ and $\mathbf{b}_{ug} \in \mathbb{R}^{d}$, to yield question embedding $\mathbf{u} \in \mathbb{R}^d$ and answer embedding $\mathbf{g} \in \mathbb{R}^{5 \times d}$ where $d$ denotes the memory dimension.

Video representation $\mathbf{v}$ and subtitle representation $\mathbf{s}$ are embedded independently to generate video memory $\mathbf{M}^v$ and subtitle memory $\mathbf{M}^s$. This dual memory structure enables pinpointing different temporal parts for each modality. To reflect the observation that the adjacent video clips often have strong correlations, we utilized the average pooling (Avg.Pool) layer to store the adjacent representations into a single memory slot.

As the first step of dual memory embedding, feed-forward neural network (FFN) composed of two linear FC layers with ReLU non-linearity in between is applied to embed video and subtitle representation. This operates on every element of $\mathbf{v}$ and $\mathbf{s}$ independently. Then, average pooling layer is applied to model neighboring representations together, forming video memory $\mathbf{M}^v$ and subtitle memory $\mathbf{M}^s$, i.e. \textit{dual memory}:
\begin{eqnarray}
\text{FFN}(\mathbf{x}) &=& \text{ReLU}(\mathbf{x}\mathbf{W}_1 + \mathbf{b}_1)\mathbf{W}_2 + \mathbf{b}_2, \label{eqn:2} \\
\mathbf{M}^v &=& \text{Avg.Pool}(\text{FFN}(\mathbf{v}); \theta_p, \theta_s), \label{eqn:3} \\
\mathbf{M}^s &=& \text{Avg.Pool}(\text{FFN}(\mathbf{s}); \theta_p, \theta_s), \label{eqn:4}
\end{eqnarray}
where $\theta_p$ and $\theta_s$ denotes the size and stride of the pooling, $\mathbf{x}$ indicates the each input and $\mathbf{W}$, $\mathbf{b}$ denotes the weight and bias of feed-forward neural network. Finally, generated video and subtitle memory are $\mathbf{M}^v, \mathbf{M}^s \in \mathbb{R}^{N \times d}$ where $N = \lceil T/ \theta_s \rceil$.


\subsection{Progressive Attention Mechanism}
\label{ssec:3.3}
The progressive attention mechanism in Fig. \ref{fig:overall}(b) takes dual memory $\mathbf{M}^v, \mathbf{M}^s$, question embedding $\mathbf{u}$ and answer embedding $\mathbf{g}$ as inputs, and progressively attends and updates the dual memory. While iteratively taking question and answers for temporal attention generation, memories are progressively updated to accumulate cues to locate relevant temporal parts for answering the question. We observed that single-step temporal attention on memory networks \cite{MovieQA,rwmn} often generates blurry attention map.	The multi-step nature of progressive attention mechanism enables generating sharper attention distribution. Unnecessary information from memory is filtered out at each iteration. 

The first step of progressive attention mechanism is temporal attention by question embedding $\mathbf{u}$. The attention weights are obtained by calculating the cosine similarity between each memory slot and question embedding $\mathbf{u}$ as in Eqs.\ref{eqn:5}, \ref{eqn:6}. The dual memory is multiplied by the attention weights and is followed by linear FC layer to be updated as in Eqs.\ref{eqn:7}, \ref{eqn:8}. The attention operates independently for the video memory $\mathbf{M}^v$ and the subtitle memory $\mathbf{M}^s$:
\begin{eqnarray}
\alpha_{Mu}^v &=& \text{softmax}(\mathbf{u} \mathbf{M}^{v\intercal}), \label{eqn:5} \\
\alpha_{Mu}^s &=& \text{softmax}(\mathbf{u} \mathbf{M}^{s\intercal}), \label{eqn:6}  \\
\mathbf{M}^v &\leftarrow& (\alpha_{Mu}^v \odot \mathbf{M}^v)\mathbf{W}_{Mu}^v + \mathbf{b}_{Mu}^v, \label{eqn:7}\\
\mathbf{M}^s &\leftarrow& (\alpha_{Mu}^s \odot \mathbf{M}^s)\mathbf{W}_{Mu}^s + \mathbf{b}_{Mu}^s, \label{eqn:8}
\end{eqnarray}
where $\alpha_{Mu}^v, \alpha_{Mu}^s \in \mathbb{R}^{N}$ denote the temporal attention weight for $\mathbf{M}^v, \mathbf{M}^s$, respectively. The learnable parameters for linear FC layer is denoted by $\mathbf{W}_{Mu}, \mathbf{b}_{Mu}$ , $\leftarrow$ indicates the update operation and $\odot$ represents the element-wise multiplication with broadcasting on appropriate axis.

The second step of progressive attention mechanism is temporal attention by answers. This step is similar to the first step except it utilizes answer embedding $\mathbf{g}$ to attend updated dual memory $\mathbf{M}^v, \mathbf{M}^s$:
\begin{eqnarray}
\alpha_{Mg}^v &=& \text{softmax}(\mathbf{g} \mathbf{M}^{v\intercal}), \label{eqn:9} \\
\alpha_{Mg}^s &=& \text{softmax}(\mathbf{g} \mathbf{M}^{s\intercal}), \label{eqn:10} \\
\mathbf{M}^v &\leftarrow& (\alpha_{Mg}^v \odot \mathbf{M}^v)\mathbf{W}_{Mg}^v + \mathbf{b}_{Mg}^v, \label{eqn:11} \\
\mathbf{M}^s &\leftarrow& (\alpha_{Mg}^s \odot \mathbf{M}^s)\mathbf{W}_{Mg}^s + \mathbf{b}_{Mg}^s, \label{eqn:12}
\end{eqnarray}
where $\alpha_{Mg}^v$ and $\alpha_{Mg}^s \in \mathbb{R}^{5 \times N}$ denote the temporal attention weights for dual memory and $\mathbf{M}^v, \mathbf{M}^s \in \mathbb{R}^{5 \times N \times d}$ represent the updated video and subtitle memory, respectively.

\textbf{Multiple Hops Extension.}
As described above, the progressive attention mechanism attends the dual memory only once for each attention step. In this case, the dual memory may contain much irrelevant information and lack capability to query complicated semantics to answer the question. Progressive attention can be naturally extended to utilize multiple hops \cite{e2ememory} for fine-grained extraction of abstract concepts and reasoning of high-level semantics.

Different from the memory network \cite{e2ememory} that utilizes the sum of the output $o_k$ and query $u_k$ of $k$-th hop as the query of next hop, we use the same question embedding $\mathbf{u}$ with updated dual memory $\mathbf{M}^{(k)}$ for $k$-th hop. Each attention step in Eqs. \ref{eqn:5}-\ref{eqn:8}, \ref{eqn:9}-\ref{eqn:12} is repeated $h_{Mu}, h_{Mg}$ times, respectively. Each attend and update operations can be expressed as:
\begin{eqnarray}
\alpha^{(k)} &=& \text{softmax}(\mathbf{x} \mathbf{M}^{(k-1)\intercal}), \label{eqn:13} \\
\mathbf{M}^{(k)} &\leftarrow& (\alpha^{(k)} \odot \mathbf{M}^{(k-1)})\mathbf{W}^{(k)} + \mathbf{b}^{(k)}, \label{eqn:14}
\end{eqnarray}
where the subscripts and superscripts corresponding to each equation are omitted to avoid repetition, and $\mathbf{x}$ indicates $\mathbf{u}$ or $\mathbf{g}$ for each step of progressive attention.

\subsection{Dynamic Modality Fusion}
Dynamic modality fusion in Fig. \ref{fig:overall}(c) aggregates dual memory into fused output $\mathbf{o}$ at the end of each progressive attention step. Different question requires different modality to infer the answer. Consider the question \textit{"What drink bottle is at the table when Robin, Lily, Marshall and Ted are talking to each other?"}. In this case, the video modality would be more important than subtitle modality. Similar to modality attention \cite{Hori_2017_ICCV,Kang_2018_ECCV}, dynamic modality fusion is soft-attention based algorithm that determines the contribution of each modality for answering the question. 

Given dual memory $\mathbf{M}^v, \mathbf{M}^s$, dynamic modality fusion first sum each memory along temporal axis and compute cosine similarity with question embedding $\mathbf{u}$ to calculate attention score.:
\begin{eqnarray}
\mathbf{o}^m &=& \sum_{n=1}^{N}\mathbf{M}^m, \label{eqn:15}\\
\alpha_{\text{DMF}} &=& \text{softmax}(\mathbf{u}[\mathbf{o}^{v};\mathbf{o}^{s}]^{\intercal}), \label{eqn:16}
\end{eqnarray}
where $m$ indicates each modality $\mathbf{v}$ or $\mathbf{s}$, $\mathbf{o}^m$ represents the output of each memory, $N$ denotes the temporal length of dual memory, and $\alpha_{\text{DMF}}$ denotes attention weights. Finally, the fused output $\mathbf{o}$ is computed by weighted summing between attention weight and memory output:
\begin{equation}
\mathbf{o} = \sum_m \alpha_{\text{DMF}}^m \mathbf{o}^m.
\end{equation}
The learned attention weight can be interpreted as contribution or importance of each modality on answering the question. By regulating the ratio of each modality on fused output, dynamic modality fusion leads to stable learning by discarding information from unnecessary modality.

\subsection{Belief Correction Answering Scheme}

Belief correction answering scheme in Fig. \ref{fig:overall}(d) selects the correct answer among five candidate answers. Rather than determining the prediction score once, belief correction answering scheme successively corrects prediction score by observing diverse source of information. This mimics the multi-step reasoning process of human answering difficult questions \cite{human}. Combined with progressive attention and dynamic modality fusion, this multi-step reasoning approach of PAMN strengthens the model's ability to extract high-level meaning from the multimodal data. 

Belief $\mathbf{B} \in \mathbb{R}^5$ denotes the prediction score on the candidate answers. The prediction probability $\mathbf{z} \in \mathbb{R}^{5}$ is computed by normalizing the belief, and the answer $y$ is predicted with the highest probability:
\begin{eqnarray}
\mathbf{z} &=& \text{softmax}(\mathbf{B}), \\
y &=& \text{arg max}_{i \in [5]}(\mathbf{z}_i).
\end{eqnarray}
One way of initializing belief would be \textit{null initialization} that endows all candidate answers with equal probabilities before observing any information. To reflect this unbiased initialization, the belief $\mathbf{B}$ is initialized as zero vector.

Belief correction answering scheme adopts three-step belief correction; \textit{u}-, \textit{Mu}- and \textit{Mg}-correction. For each correction step, the belief is corrected by accumulating the similarity between answer embedding $\mathbf{g}$ and the observed information. Belief is first corrected by only considering the question, i.e. \textit{u}-correction. The intuition is that human often builds prior biases after skimming through only the question and candidate answers:
\begin{eqnarray}
&&\mathbf{B}_{u} = \mathbf{u}\mathbf{g}^\intercal, \\
&&\mathbf{B} \leftarrow \mathbf{B} + \mathbf{B}_{u}.
\end{eqnarray}
Then for \textit{Mu}- and \textit{Mg}-correction, the outputs of first and second progressive attention steps, $\mathbf{o}_{Mu}$ and $\mathbf{o}_{Mg}$, are considered. Again, the similarities between answer embedding $\mathbf{g}$ are computed:
\begin{eqnarray}
&&\mathbf{B}_{Mu} = \mathbf{o}_{Mu} \mathbf{g}^\intercal,  \\
&&\mathbf{B}_{Mg, i} = \mathbf{o}_{Mg, i}\mathbf{g}_{i}^\intercal.
\end{eqnarray}
Finally, the belief is corrected to infer correct answer:
\begin{eqnarray}
&&\mathbf{B} \leftarrow \mathbf{B} + \beta_{Mu}\mathbf{B}_{Mu}, \\
&&\mathbf{B} \leftarrow \mathbf{B} + \beta_{Mg}\mathbf{B}_{Mg},
\end{eqnarray}
where the correction weights $\beta_{Mu}$, $\beta_{Mg}$ are hyper parameters that scales corresponding belief correction. Note that the belief is normalized to have unit norm after each correction. 

\section{Experiments}
\label{sec:4}

\subsection{Dataset}
\label{ssec:4.1}

\textbf{MovieQA} \cite{MovieQA} benchmark is constructed for movie story QA which consists various sources of information such as movie clip, subtitle, plot synopses, scripts and DVS transcriptions. MovieQA dataset contains 408 movies with corresponding 14,944 multiple-choice questions. MovieQA benchmark consists of 6 tasks according to which sources to be used. This paper focuses on video+subtitles task which is the only task utilizing movie clip. Since only 140 movies contain video clips, there are 6,462 question-answer pairs which splits into 4,318 training, 886 validation and 1,258 test samples.

\textbf{TVQA} \cite{lei2018tvqa} benchmark is video story QA dataset on TV show domain. It consists of total 152.5k question-answer pairs on six TV shows: \textit{The Big Bang Theory, How I Met Your Mother, Friends, Grey's Anatomy, House, Castle}. Each split of TVQA contains 122k, 15.25k, 15.25k for train, validation and test, respectively. Unlike MovieQA which considers whole movie as input, TVQA contains 21,793 short clips of 60/90 seconds segmented from the original TV show for question-answering.

\subsection{Feature extraction}
\label{ssec:4.2}
For fair comparison, we extracted visual and textual features similar to previous works \cite{rwmn,MovieQA} and fixed them during training. 

\textbf{Textual feature} Each sentence from question, candidate answers and subtitle are divided into sequence of words, then each word is embedded by skip-gram model \cite{word2vec} provided by Tapaswi \textit{et al.} \cite{MovieQA} which is trained on MovieQA plot synopses. In order to encode the order of words within a sentence, position encoding (PE) \cite{e2ememory} is utilized to obtain textual feature. For example in the case of question, $\mathbf{q} = \sum_{n}\mbox{PE}(\mathbf{q}_n) \in \mathbb{R}^{300}$ where each $\mathbf{q}_n$ indicates word vector.

\textbf{Visual feature} Movies are divided into video clips that are temporally aligned with each sentence of the subtitle. The frames are sampled from each video clip with the rate of 1 fps. Then, frame feature of size 1536 is extracted from \textquotedblleft Average Pooling\textquotedblright\ layer on Inception-v4 \cite{inception}. Finally, mean-pooling over all frame features from the corresponding video clip produces the visual feature, $\mathbf{v}_i \in \mathbb{R}^{1536}$.

\begin{table}[t]
	\begin{center}
		\begin{tabular}{l||c|c}
			\hline
			Methods                        & valid Acc. & test Acc.    \\ 
			\hline \hline
			SSCB w/o Sub                   & 21.60 & -                 \\
			SSCB w/o Vid                   & 22.30 & -                 \\
			SSCB \cite{MovieQA}            & 21.90 & -                 \\
			MemN2N w/o Sub                 & 23.10 & -                 \\
			MemN2N w/o Vid                 & 38.00 & -                 \\
			MemN2N \cite{MovieQA}          & 34.20 & -                 \\
			DEMN \cite{demn}               & 44.70 & 29.97             \\
			RWMN \cite{rwmn}               & 38.67 & 36.25             \\
			FVTA \cite{fvta}               & 41.00 & 37.30             \\
			LMN \cite{lmn}                 & 42.50 & 39.03             \\
			MDAM \cite{mdam}               & - & 41.41             \\
			\hline
			PAMN w/o Sub                   & 42.33 & -                 \\
			PAMN w/o Vid                   & 42.56 & -                 \\
			PAMN                           & 43.34 & \textbf{42.53}    \\
			\hline
		\end{tabular}
	\end{center}
	\caption{Accuracy comparison on the validation and test set of MovieQA benchmark of Video+Subtitles task. PAMN achieves the state-of-the-art performance. The test set accuracy is obtained from online evaluation server. And \textquoteleft-\textquoteright\ indicates that the performance is not provided.}
	\label{tab:acc}
\end{table}

\begin{table}[t]
	\begin{center}
		\begin{tabular}{l||c|c}
			\hline
			Methods                        & Video Feat. & test Acc.      \\ 
			\hline \hline
			Longest Answer                 & -           & 30.41          \\ 
			\hline
			TVQA \cite{lei2018tvqa}        & img         & 63.57          \\
			TVQA \cite{lei2018tvqa}        & reg         & 63.19          \\
			TVQA \cite{lei2018tvqa}        & cpt         & 65.46          \\
			\hline
			PAMN                           & img         & 64.61          \\
			PAMN                           & cpt         & \textbf{66.77} \\
			\hline
		\end{tabular}
	\end{center}
	\caption{Accuracy comparison on the test set of TVQA benchmark without timestamp annotation. We utilized the video and text features extracted by Lei \textit{et al.} \cite{lei2018tvqa}.}
	\label{tab:acc_tvqa}
\end{table}
\subsection{Implementation details}
\label{ssec:4.3}

The entire architecture was implemented using Tensorflow~\cite{abadi2016tensorflow} framework. All the results reported in this paper were obtained using the Adagrad optimizer~\cite{duchi2011adaptive} with a mini-batch size of 32 and the learning rate of 0.001. All the experiments were performed under CUDA acceleration with single NVIDIA TITAN Xp (12GB of memory) GPU. In all the experiments, the recommended train / validation / test split was strictly observed.

\subsection{Quantitative Results}
\label{ssec:4.4}

Table \ref{tab:acc} compares the validation and test accuracy on the MovieQA benchmark of Video+Subtitles task. We compare the performance of PAMN with other state-of-the-art architecture. The ground-truth answers for MovieQA test set are not observable and the evaluation on the test set can only be performed once every 72 hours through an online evaluation. On MovieQA benchmark, PAMN exhibits the state-of-the-art results by attaining test accuracy of 42.53\%. It outperforms the runner-up, MDAM \cite{mdam} (41.41\%) by 1.12\% and the third place, LMN \cite{lmn} (39.03\%) by 3.50\%. Note the MDAM is an ensemble of 20 different models, while PAMN is a single model. 

In order to evaluate the effectiveness of each modality, experiments based on using only video and subtitle were also conducted: PAMN w/o Sub and PAMN w/o Vid. From near random-guess performances of SSCB w/o Sub \cite{MovieQA} and MemN2N w/o Sub \cite{MovieQA} as shown in Table. \ref{tab:acc}, it is noticed that movie story understanding is difficult using only video. The PAMN w/o Sub attains large performance gain of 19.23\% compared to MemN2N w/o Sub. It even achieves performance comparable to LMN \cite{lmn} which exploits both video and subtitle. PAMN understands movie story even without observing subtitle. From Table.\ref{tab:acc}, it is noticed that PAMN performs better than PAMN w/o Vid and PAMN w/o Sub which indicates both video and subtitle provides conducive information in improving prediction. 

Table \ref{tab:acc_tvqa} shows performance comparison on TVQA benchmark without timestamp annotation. In this experiment, we utilized the video and text features extracted by Lei \textit{et al.} \cite{lei2018tvqa} (i.e. ImageNet and visual concept feature for video and GloVe feature for text) for fair comparison. Further, we encoded the sentence feature using LSTM instead of position encoding. On TVQA benchmark, PAMN outperforms state-of-the-art result by attaining test accuracy of 66.77\% with visual concept feature.

\subsection{Ablation Study}
\label{ssec:4.5}

\begin{table}[t]
	\begin{center}
		\begin{tabular}{l||c|c}
			\hline 
			Methods                                   & valid Acc.     & $\Delta$  \\ \hline \hline
			PAMN w/o PA                               & 42.03          & -1.31\%   \\
			PAMN w/o Multiple Hop                     & 42.67          & -0.67\%   \\ \hline
			PAMN w/o DMF                              & 42.09          & -1.25\%   \\
			PAMN w/ MCB \cite{mcb}                    & 42.89          & -0.45\%   \\
			PAMN w/ MFB \cite{mfb}                    & 42.55          & -0.79\%   \\
			PAMN w/ Tucker \cite{mutan}               & 42.89          & -0.45\%   \\ \hline
			PAMN w/o \textit{Mu,Mg}-correction        & 39.50          & -3.84\%   \\
			PAMN w/o \textit{Mg}-correction           & 41.76          & -1.58\%   \\ 
			PAMN w/o \textit{Mu}-correction           & 40.86          & -2.48\%   \\ \hline
			PAMN                             & \textbf{43.34} & -         \\ \hline
		\end{tabular}
	\end{center}
	\caption{Ablation studies of the proposed PAMN on the validation set of MovieQA benchmark. The last column shows the performance drop.}
	\label{tab:abl}
\end{table}

Table. \ref{tab:abl} summarizes the ablation analysis of PAMN on the validation set of MovieQA benchmark in order to measure the validity of the key components of PAMN. To measure to effectiveness of progressive attention mechanism, 
each temporal attention step of PAMN w/o PA utilizes dual memory obtained in Eqs. \ref{eqn:3},\ref{eqn:4}, i.e. PAMN w/o PA do not accumulate cues and each attention step operates in a parallel manner. PAMN w/o Multiple Hop attends dual memory only once for each temporal attention step. As shown in the first block of Table. \ref{tab:abl}, PAMN w/o PA underperforms PAMN, which shows that the attention accumulation by progressive attention mechanism is important in understanding movie story. Multiple hops extension is also crucial in attaining the best possible performance. For ablating dynamic modality fusion, we experiment with four variants: PAMN w/o DMF take the mean of the outputs of dual memory $\mathbf{o}^v$, $\mathbf{o}^s$, PAMN w/ MCB, MFB, Tucker use MCB \cite{mcb}, MFB \cite{mfb}, Tucker decomposition \cite{mutan,Tucker1966} instead of dynamic modality fusion, respectively. As shown in the second block of Table. \ref{tab:abl}, fusing modalities by averaging or bilinear operations show lower performance than dynamic modality fusion. This implies that question dependent modality weighting (i.e. dynamic modality fusion) helps strengthens conducive modality. To measure the effectiveness of belief correction answering scheme, the third block of Table. \ref{tab:abl} shows the experimental results of three variants: PAMN w/o \textit{Mu,Mg}-correction, PAMN w/o \textit{Mg}-correction, and PAMN w/o \textit{Mu}-correction. It is noteworthy that only using QA pairs shows much higher performance that the random baseline of 20\%. Considering $Mu$- and $Mg$-correction, PAMN w/o \textit{Mg}-correction shows 2.26\% and PAMN w/o \textit{Mu}-correction shows 1.36\% performance boosts, respectively.

\begin{table}[t]
	\begin{center}
		\begin{tabular}{c c|c c|c c|c}
			\hline
			\multicolumn{2}{c|}{\# hops} & \multicolumn{2}{|c|}{Avg. Pool} & \multicolumn{2}{|c|}{Correction} & Acc.    \\
			$h_{Mu}$     & $h_{Mg}$     & $\theta_p$           & $\theta_s$           & $\beta_{Mu}$   & $\beta_{Mg}$  &        \\ \hline \hline
			1            & 1            & 1             & 1             & 1              & 0.5           & 38.94  \\
			1            & 1            & 12            & 8             & 1              & 0.5           & 40.18  \\ \hdashline
			1            & 1            & 24            & 16            & 0.5            & 0.5           & 40.07  \\
			1            & 1            & 24            & 16            & 1              & 0.1           & 42.10   \\
			1            & 1            & 24            & 16            & 1              & 0.5           & 42.67 \\ \hdashline
			1            & 1            & 40            & 30            & 0.5            & 0.5           & 40.97 \\
			1            & 1            & 40            & 30            & 1              & 0.1           & 42.66 \\
			1            & 1            & 40            & 30            & 1              & 0.5           & 42.55 \\ \hdashline
			1            & 1            & 80            & 60            & 1              & 0.5           & 41.20        \\ \hline
			2            & 2            & 24            & 16            & 1              & 0.5           & \textbf{43.34} \\
			2            & 2            & 40            & 30            & 1              & 0.1           & 42.89          \\ \hline
			3            & 3            & 24            & 16            & 1              & 0.5           & 42.55          \\
			3            & 3            & 40            & 30            & 1              & 0.1           & 42.77          \\ \hline
		\end{tabular}
	\end{center}
	\caption{Performance variation of PAMN on the validation set of MovieQA benchmark depending on three sets of hyper parameters. $h_{Mu}, h_{Mg}$: the number of hops for attention by question $\mathbf{u}$ and answer $\mathbf{g}$, $\theta_p, \theta_s$: size and stride of Avg. Pool layer, and $\beta_{Mu}, \beta_{Mg}$: correction weights for belief correction module.}
	\label{tab:abl_2}
\end{table}

Table. \ref{tab:abl_2} summarizes the performance variation depending on three sets of hyperparameters; the number of hops for attention by question $\mathbf{u}$ and answer $\mathbf{g}$, $\theta_p, \theta_s$: size and stride of Avg. Pool layer, and $\beta_{Mu}, \beta_{Mg}$: correction weights for belief correction module. The multiple hops extension with 2-repetitions exhibits the best validation performance for PAMN. The multiple hops extension with more than three repetitions may suffer from overfitting due to the small size of dataset. The performance is positively affected by increasing $\theta_p$ and $\theta_s$, but it degrades for large $\theta_p$ and $\theta_s$ due to information blurring of Avg. Pool. 
We observed that there is no best-performing optimal correction weights. If the question representation $\mathbf{u}$ has enough information about where in the movie to focus on, $\beta_{Mu}$ should be higher, and vice versa. Furthermore, it is preferable to have smaller $\beta_{Mg}$ than $\beta_{Mu}$ since large value of $\beta_{Mg}$ dilates the effect of ug and Mu correction since the normalization is applied in between every belief correction. 

\subsection{Qualitative analysis}
\label{ssec:4.6}

\begin{figure*}[t!]
	\begin{center}
		\includegraphics[width=\textwidth]{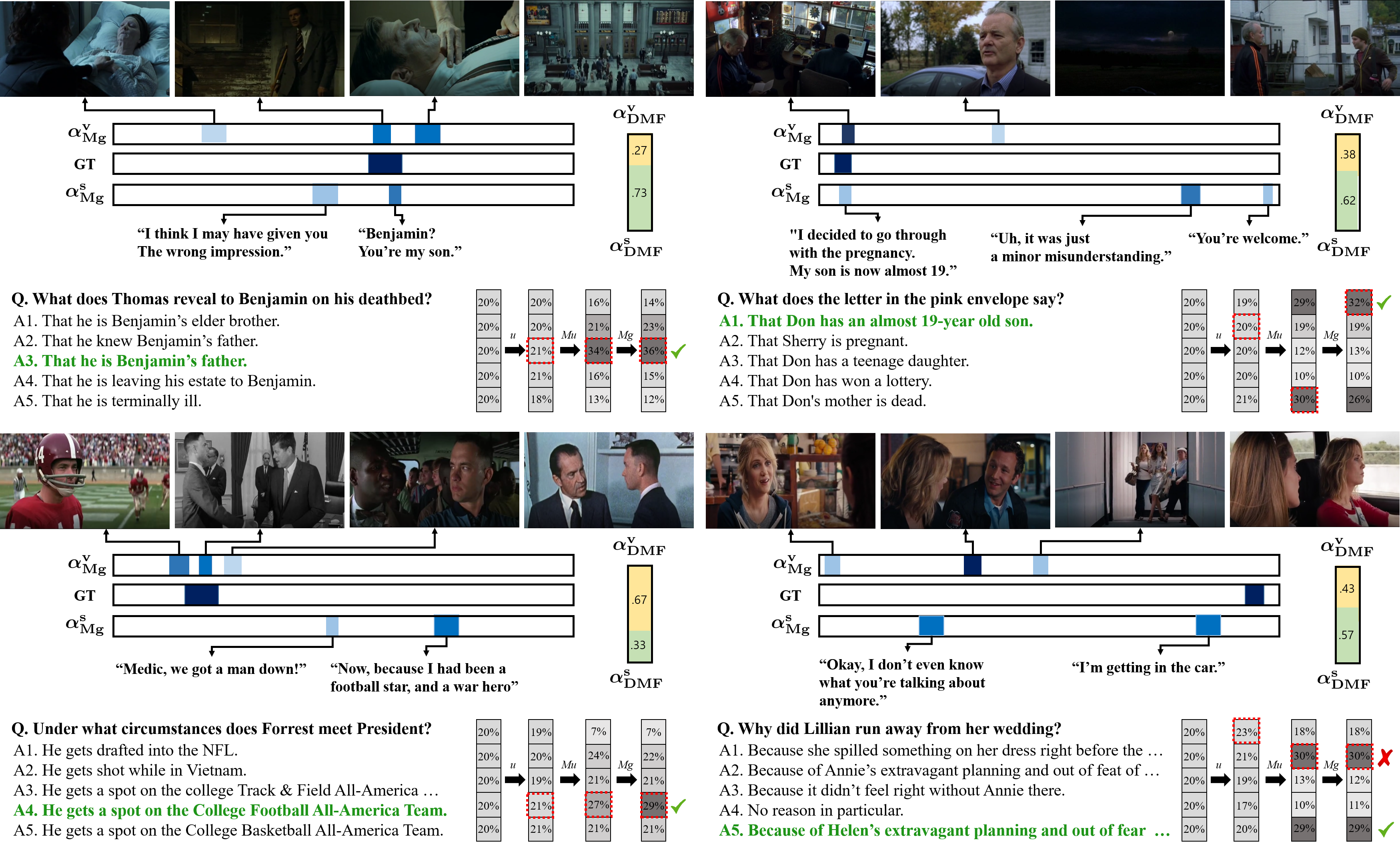}
		\caption{Qualitative examples of MovieQA benchmark solved by PAMN (the last example is failure case). Green sentences and check symbols indicate correct answers and red dotted boxes highlight PAMN's prediction at each belief correction step. For failure cases, red \textquoteleft x\textquoteright\ symbols indicate the the incorrect selection. $\alpha_{Mg}^v, \alpha_{Mg}^s$ represents temporal attention obtained by progressive attention mechanism, $\alpha_{\text{DMf}}^v, \alpha_{\text{DMF}}^s$ denotes attention obtained by dynamic modality fusion. The temporal attention by PAMN matches well with groundtruth (GT) where the question is generated. Observing diverse source of information, PAMN successfully corrects the belief toward correct answer.}
		\label{fig:qual}
	\end{center}
\end{figure*}
\begin{figure}[h!]
	\begin{center}
		\includegraphics[width=0.85\linewidth]{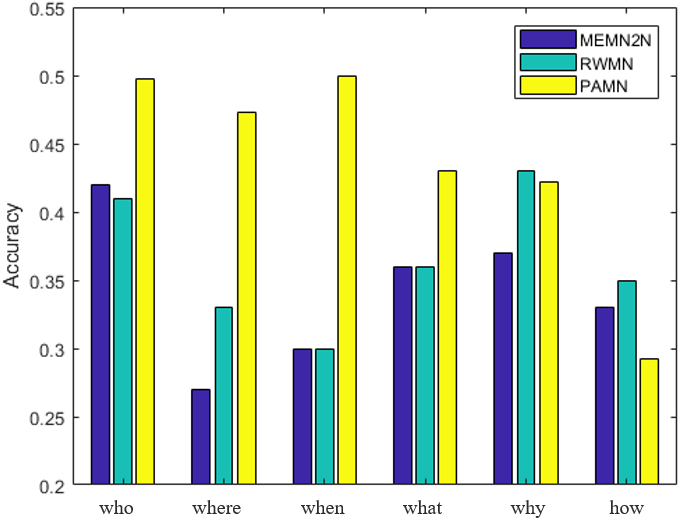}
		\caption{Accuracy comparison with respect to the first word of the question between MemN2N \cite{MovieQA}, RWMN \cite{rwmn} and PAMN on the validation set of MovieQA. PAMN outperforms on the majority of the question types.}
		\label{fig:qual_2}
	\end{center}
\end{figure}

The Fig. \ref{fig:qual} illustrates the selected qualitative examples of PAMN. Each example provides the temporal attention map $\alpha_{Mg}^v, \alpha_{Mg}^s$ from progressive attention mechanism, the ground-truth (GT) temporal part where the question was generated from, the attention weights $\alpha_{DMF}^v, \alpha_{DMF}^s$ from dynamic modality fusion, and the inference path of belief correction answering scheme. The generated temporal attention well matches with the GT which indicates that PAMN successively learns where to attend. The weights $\alpha_{\text{DMF}}^v, \alpha_{\text{DMF}}^s$ adaptively scales depending on the question type which implies that PAMN learns what modality to use without additional supervision. For some cases, PAMN predicts the correct answer at the \textit{u}-correction step while for other cases the correct answer is determined at the last (\textit{Mg}) step. PAMN is an interpretable architecture in that the inference path and the attention map provide the trace of where PAMN attends and what information source it used to answer the question.

The Fig. \ref{fig:qual_2} exhibits the accuracy comparison with respect to the first word of the question between MemN2N \cite{MovieQA}, RWMN \cite{rwmn} and PAMN on the validation set of MovieQA benchmark. The results on 5W1H question types: \textit{Who, Where, When, What, Why} and \textit{How} are analyzed. Typically, answering \textit{who, where, when, what} questions require pinpointing the temporal parts relevant to the question (e.g., When do the loyalists take over Air Force One?, What does Korshunov demand from Vice President Bennett?). On the other hand, answering \textit{why, how} questions require understanding the contextual information over the whole movie (e.g., How do Schmidt and Jenko's fake identities end up getting switched?, Why does Mozart's financial situation get worse and worse?). We observed that PAMN outperforms MemN2N and RWMN on the majority of question types. Especially, PAMN attains 20\% and 13\% performance boosts on \textit{when, where} questions, respectively which  implies the superiority of PAMN to pinpoint the movie story.

\section{Conclusion}
\label{sec:5}

In this paper, a movie story question answering (QA) architecture referred to as Progressive Attention Memory Network (PAMN) was proposed. The main challenges of movie story QA were summarized as: (1) pinpointing the temporal parts relevant to answer the question is difficult (2) different questions require different modality to infer the answer. Proposed PAMN make use of three main features to tackle aforementioned challenges: (1) progressive attention mechanism, (2) dynamic modality fusion and (3) belief correction answering scheme. We empirically demonstrated that proposed PAMN is valid by showing the state-of-the-art performance on MovieQA and TVQA benchmark dataset.

{\small
\bibliographystyle{ieee_fullname}
\bibliography{egbib}
}

\end{document}